\documentclass[sigconf]{acmart}
\AtBeginDocument{%
  \providecommand\BibTeX{{%
    \normalfont B\kern-0.5em{\scshape i\kern-0.25em b}\kern-0.8em\TeX}}}

\setcopyright{acmcopyright}
\copyrightyear{2022}
\acmYear{2022}
\acmDOI{}

\acmConference[KDD-UC'22]{KDD Undergraduate Consortium}{Aug 14--18,2022}{Washington, DC, USA}
\acmPrice{}
\acmISBN{}

\usepackage{adjustbox}
\usepackage{caption}
\usepackage{booktabs}
\begin{document}
\setlength\belowcaptionskip{-5pt}
\title{EEG4Students: An Experimental Design for EEG Data Collection and Machine Learning Analysis}

\author{Zheng Zhou}
\authornote{Both authors contributed equally to this research.}
\email{zhengzhou@brandeis.edu}
\orcid{0000-0001-9313-2106}
\affiliation{%
  \institution{Brandeis University}
  \streetaddress{415 South Street}
  \city{Waltham}
  \state{Massachusetts}
  \country{USA}
  \postcode{02453}
}

\author{Guangyao Dou}
\email{guangyaodou@brandeis.edu}
\orcid{0000-0001-8011-9658}
\authornotemark[1]
\affiliation{%
  \institution{Brandeis University}
  \streetaddress{415 South Street}
  \city{Waltham}
  \state{Massachusetts}
  \country{USA}
  \postcode{02453}}

\renewcommand{\shortauthors}{Zhou and Dou.}

\begin{abstract}
Using Machine Learning and Deep Learning to predict cognitive tasks from electroencephalography (EEG) signals has been a fast-developing area in Brain-Computer Interfaces (BCI). However, during the COVID-19 pandemic, data collection and analysis could be more challenging. The remote experiment during the pandemic yields several challenges, and we discuss the possible solutions. This paper explores machine learning algorithms that can run efficiently on personal computers for BCI classification tasks. The results show that Random Forest and RBF SVM perform well for EEG classification tasks. Furthermore, we investigate how to conduct such BCI experiments using affordable consumer-grade devices to collect EEG-based BCI data. In addition, we have developed the data collection protocol, EEG4Students, that grants non-experts who are interested in a guideline for such data collection. Our code and data can be found at \url{https://github.com/GuangyaoDou/EEG4Students}.
\end{abstract}



\maketitle

\section{Introduction}
Researchers from Computer Science, Neuroscience, and Medical fields have applied EEG-based Brain-Computer Interaction (BCI) techniques in many different ways \cite{lotte2018review,craik2019deep,qu2020identifying,appriou2020modern,lotte2010regularizing, lotte2014tutorial}, such as diagnosis of abnormal states, evaluating the effect of the treatments, seizure detection, motor imagery tasks \cite{devlaminck2010circular,lotte2015signal,lotte2015towards,bashivan2014spectrotemporal,bashivan2015learning,bashivan2016mental}, and developing BCI-based games \cite{coyle2013guest}. Previous studies have demonstrated the great potential of machine learning, deep learning, and transfer learning algorithms \cite{lotte2007review,zhang2020survey,zhao2019bira,roy2019deep,miller2019library,kaya2018large,zhao2020sea,lotte2018bci,bird2018study, zhao2021mt,an2020transfer} in clinical data analysis.

However, EEG datasets' size is still relatively small compared with peers in Computer Vision and Natural Language Processing \cite{lotte2018review,qu2018eeg,craik2019deep}. The COVID-19 pandemic makes it harder to conduct BCI experiments in person at the research labs. Also, EEG signals have noise issues, mainly because of the contact of sensors and skin for several current non-invasive consumer-grade devices. The outlier issue is also a concern for the EEG data because of the difficulties subjects have in concentrating on the experimental tasks during the entire session. Although the current debate on whether the clinical or consumer-grade device is more appropriate for research remains, there is no doubt that more affordable products such as the euroSky, Muse, and OpenBCI are showing promising results with validation and repeatability\cite{sawangjai2019consumer}. Therefore, we see this opportunity to utilize the more accessible consumer-grade devices to enlarge the current dataset. Given the simplicity to use, we can also improve the data collection efficiency. 

On the other hand, current algorithms and frameworks are more for in-person clinical experiments and less for the possible virtual experiment during the pandemic or with non-expert subjects. Given the need for a larger dataset in the field, our \textbf{research questions} are: 
1. Can we contribute to the current BCI dataset with data from a large non-expert subject population? 
2. Can we design easy-to-use BCI experiments to collect EEG data with consumer-grade devices for students?
3. Can we develop a step-by-step guide for such a data collection and machine learning analysis process so that we can still collect EEG data during the pandemic? 

This paper is organized as follows: part two elaborates on the basic experimental design of our research; part three demonstrates a detailed guideline for non-expert subjects' data collection and possible training; part four provides our data analysis workflow; part five presents the experiment result that reflects the experiment success and answers the research questions one and two; parts six and seven are the discussion of limitations along with possible future works and the conclusion, which address the research question three. 

\section{Experiment Design}

Several EEG experiments focus on college students' high-level cognitive tasks frequently conducted, as mentioned in Table \ref{tasks_experiment}.

In this experiment, we are trying to replicate the experiment from the Think-Count-Recall (TCR) paper \cite{qu2020multi} in both in-person and remote ways via Zoom, which is the classification of five cognitive tasks: Think, Count, Recall, Breathe, and Draw. Scalp-EEG signals are recorded from nineteen subjects. Each subject participates in six sessions, and each session is five minutes long. For each session, there are five tasks, and each task is one minute. Tasks are selected by the subjects and researchers based on frequent tasks in study environments for students in their everyday life. Each subject completes six sessions over several weeks.  
The programs are run on a 2018 Macbook Pro with a 2.2GHz 6-core Intel Core i7 processor and 16 GB of memory. The Python version is 3.8. The scikit-learn version is 0.24.1. 

\begin{table}[!b]
\begin{adjustbox}{width=0.999\columnwidth,center}
\begin{tabular} {cccccc}
\hline
(E) T & 1 & 2 & 3 & 4 & 5\\
\hline
\cite{qu2018personalized} & Math & Close-eye Relax & Read & Open-eye Relax & None \\
\cite{qu2018eeg} & Python Passive & Math Passive & Python Active & Math Active & None \\
\cite{qu2020using} & Read & Write Copy & Write Answer & Type Copy & Type Answer \\
\cite{qu2020multi} & Think & Count & Recall & Breathe & Draw\\
\hline
\end{tabular}
\end{adjustbox}
\vspace{.1in}
\caption{\label{tasks_experiment} Tasks (T) in Experiments (E)}
\end{table}

\section{Data Collection}
 
\begin{figure}[!t]
\centering
  \includegraphics[width=0.50\textwidth]{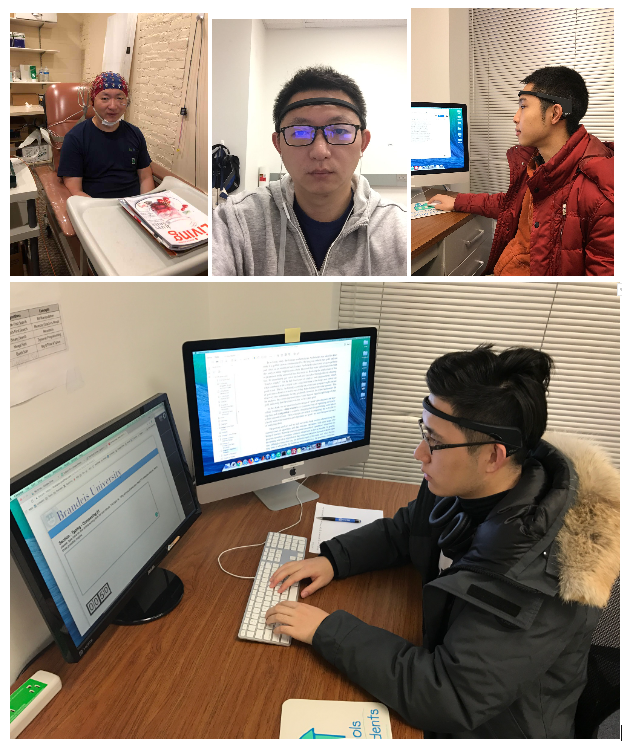}
  \caption{College Students using EEG Devices}
  \label{fig:four_in_one}
\end{figure}

As \cite{ienca2018brain,qu2018personalized,portillo2021mind,cannard2021validating} mentioned, several affordable (less than three hundred dollars) non-invasive consumer-grade EEG headsets are commercially available. As shown in Figure \ref{fig:four_in_one}, we have pilot-tested several clinical and non-clinical EEG devices with college students. The Top left cell in Figure \ref{fig:four_in_one} shows an example of the wearing of a clinical device while others demonstrate consumer-grade devices. Muse Headsets are used as an example for demonstration

\begin{figure}[!t]
\centering
  \includegraphics[width=0.999\columnwidth]{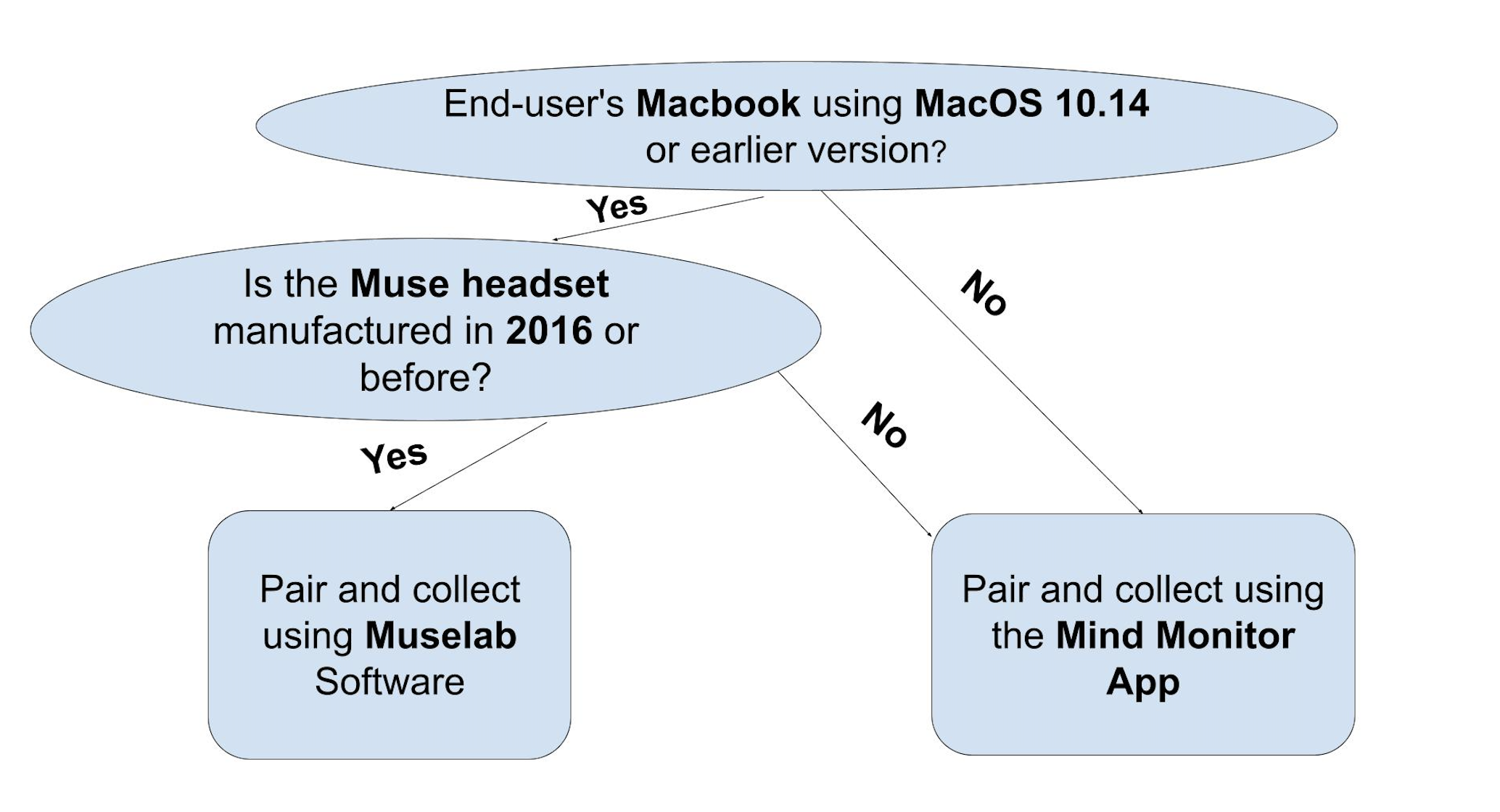}
  \caption{Data Collection App Selection}
  \label{fig:data_collection_app}
\end{figure}

\subsection{Acquiring Muse Headset Collecting Skills}
The data collecting skills are delivered and acquired entirely through in-person settings before the outbreak of COVID-19 and virtual zoom meetings for virtual participants during the pandemic. The EEG coach demonstrates the headset wearing and posture position on live video to multiple future EEG coaches simultaneously, simulating a virtual classroom setting.\textbf{As shown in Figure \ref{fig:data_collection_app}}, for end-users who have a muse headset manufactured in 2016 or before and have a computer with macOS 10.14 or former version. The MuseLab software is introduced and installed during the virtual session. Via screen-sharing, the EEG coach demonstrates and guides the trainee in step-by-step data acquisition instruction. All applications are installed from Muse Developer Resource as a Muse IO folder, including muse-io, muse-player, Muselab application, and licenses. 
For Muse headsets manufactured after 2016 or computer systems other than macOS 10.14 and the former version, a different application, "Mind Monitor," is required to collect EEG signals since the latest version headset and the Muselab software are not compatible. Mind Monitor can be directly installed from the ios APP store and paired with the muse headset following the instruction provided by the application. 

The Muse headset has four receptive fields, two near the subject's anterior side and two on the lateral sides. The headset wearing is similar to glasses-wearing, but since the headsets are not customized fitting, some participants have trouble recording consistently accurate EEG signals on the lateral receptive fields. In order to resolve the issue, rubber bands are introduced and tightened on the posterior side of the subject's head, securing the headset in place. In turn, the signal consistency is significantly improved.
  
\begin{figure}[!b]
\centering
  \includegraphics[width=0.999\columnwidth]{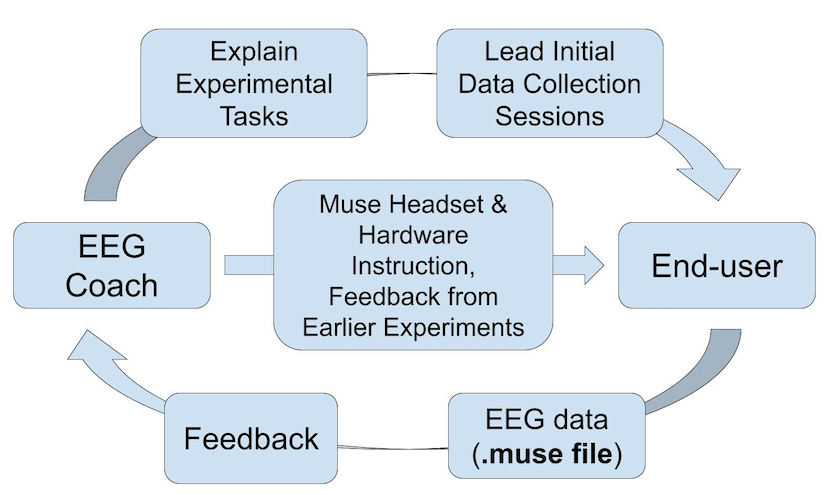}
  \caption{Data Collection Flow Chart, User and Coach}
  \label{fig:data_collection_user_coach}
\end{figure}

\subsection{Data Collection Protocol}

Two individuals, the EEG coach and the end-user are usually required to complete the data acquisition of such non-invasive EEG signals via Muse headset and Muse Recording software, As shown in Figure \ref{fig:data_collection_user_coach}. During the pandemic, data collection takes place virtually, meaning the EEG coaches have to communicate with the end-user and provide instruction through a screen-sharing virtual environment throughout the data acquisition. 

\subsubsection{\textbf{In-Person Data Collection Sessions}}

During in-person data collection sessions, an EEG coach and a test subject are located in the same room with two monitor screens. One screen monitors the headset connectivity and signals consistency, and the other displays the Qualtrics protocol to the subject. After the EEG coach successfully establishes the Muse headset connection, the participants put it on and proceed as directed by the EEG coach and the Qualtrics protocol. 

\subsubsection{\textbf{Virtual Data Collection Sessions}}

During the pandemic, all our experiments are forced to move online. Therefore, Muse headsets are mailed to participants or end-users along with instructions to establish a Bluetooth connection. When collecting data, the EEG coach would join the end-user in a zoom meeting and instruct the end-user to connect the headset and walk through the Qualtrics protocol. Zoom meeting allows the end-users to share their screen to help the EEG-coaches to monitor the data collection process and provide a remote control for necessary assistance. 

\subsubsection{\textbf{Muse Headset Usage Via MuseLab and Mind Monitor}}
Muse headset pairing can be initiated by holding the start icon on the headset for five seconds until the light indicator on the headset starts flashing, indicating ready for pairing. The EEG coach instructs the end-user to open the muse.io file on the end user's computer and connect the headset via Bluetooth. Muse.io file will automatically connect to the headset if it is ready for pairing and located in range. When the light indicator stops flashing and remains turned on, and the terminal message on the muse.io file indicates wording such as "connection successful" and displays the headset's battery life, the devices are paired. After the connection is established, the end-user is instructed to open the Muselab application and adjust the OSC port 5000 to receive the EEG signal from the headset. Before putting on the headset, the last step is to instruct the end-user to add a new visualization of EEG signals as a scrolling line graph under the visualizers' drop-down menu.

For end-users who use the MindMonitor app, the data collection protocol is similar. The end-users open the app on their compatible tablets or smartphones and a computer to follow the instructions from the EEG coach and the Qualtrics. 

After connecting the Muse headset via Bluetooth, the recording format is changed in the settings to "Muse," and the recording interval is set as "constant." The end-users are then prompted to put on the headset, and the scrolling line graphs from four Muse receptive fields will be visible on the MindMonitor app. The recording process is identical to Muslab, except that the recorded files are saved by default in plain text CSV format and are automatically uploaded to Dropbox. 

If the end-users are solely one-time participants, a screen remote control can be done to ensure efficiency. If the end-users ought to be trained as repeated users or EEG coaches, they will follow the instruction provided by the EEG coach and acquire the experiment protocol during a screen share. As soon as the EEG visualization appears on the application window, the end-user is instructed to put on the headset and open the experiment file on Quatrics, an online survey system we use to give step-by-step instructions. The end-user is asked to keep the Muselab application, and the Qualtrics file visible on the screen share to ensure the EEG coach can monitor the EEG signal receptive status by the headset while proceeding with the experiment. The end-user will follow the combined instruction by the EEG coach and the Qualtrics file to experiment. Once data collection is finished, the end-user will upload the .muse file onto the shared Google Drive for further data analysis.     

\section{Data Analysis}

Most EEG recording applications have a toolset to convert the recording files to TXT or CSV files. Afterward, we can pick the subset of data we plan to use for further analysis; we recommend starting with the absolute value of the EEG signals.
 
\begin{figure}[!b]
\centering
  \includegraphics[width=0.999\columnwidth]{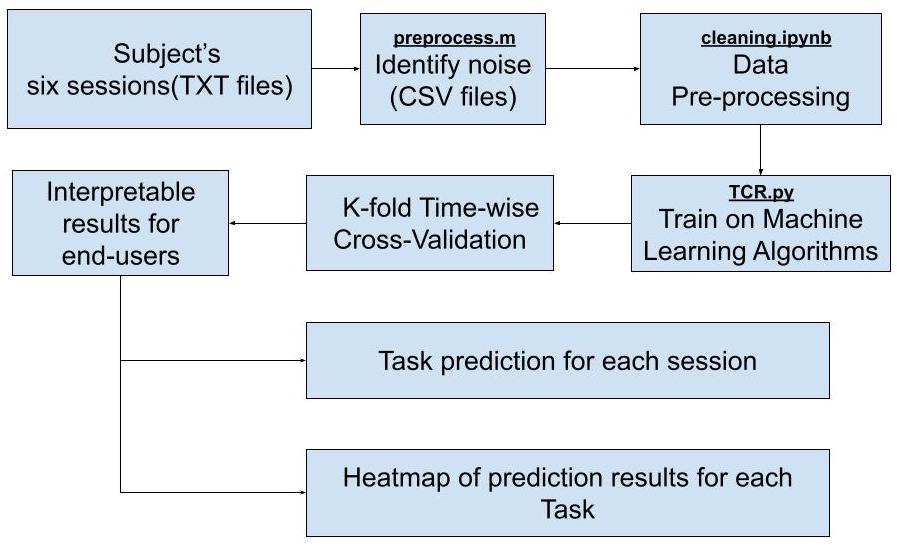}
  \caption{Data Analysis Flow Chart}
  \label{fig:EEG_Data_analysis_flow_chart}
\end{figure}

As shown in Figure \ref{fig:EEG_Data_analysis_flow_chart}, once we have six TXT files for all six data sessions, we first execute a Matlab program - preprocess.m - to identify the noises and turn all the TXT files into separate CSV files. Then, we run the cleaning.ipynb to exclude these data in the Pandas Dataframe for further analysis. Next, we execute the TCR.ipynb to train and cross-validate existing machine learning algorithms such as the Random Forest, SVM, and KNN. Lastly, the TCR.ipynb will generate a visualization of task predictions for end-users. 

\subsection{Experiments}

This section describes the subjects, data preprocessing, and validation techniques we use for experiments. 

\textbf{Subjects:} Before the pandemic, sixteen healthy subjects participated in the experiment, numbered 1 to 16. Compared to during the pandemic, we successfully collected data from only three participants remotely, and only one subject's data was adapted after denoising, numbered 17 in our experiment. We used the second trial of subject 17. Subjects who lost more than 65\% of the total data due to the data cleaning actions would be removed from the subsequent analysis. In the end, twelve subjects, including subject 17, were left to continue the analysis. Moreover, for the total of six sessions of each subject, if a session lost more than 65\% of the data during the data cleaning phase, then that session would also be excluded for further analysis. 

\textbf{Preprocessing:} As mentioned in \cite{qu2020using,qu2020multi}, for each task during each session, the first 30\% of the data, which is the first 18 seconds of each task and during the transition phase, will be removed. This has been proved reasonable during the data cleaning phase. During the EEG recording, some electrodes may have temporarily lost contact with the subjects' scalp. The result is that multiple sequential spectral snapshots from one or more electrodes have the same value. In this paper, we decide to remove such anomaly when detected for a consecutive 1.4 seconds. Such a cleaning action causes a different level of data loss for each subject. 

\textbf{Time-wise Cross-Validation:} 
We adopted time-wise cross validation\cite{qu2018eeg, saeb2016voodoo}. For each five-minute session, there were five tasks. We divided each task into seven parts, meaning each had six seconds, evenly and continuously. Then, we did seven-fold cross-validation. In each fold, we checked if each fold lost more than 65\% of the original data, then that fold was discarded for future analysis. We tested one subset in each fold and trained on the left-out subset. 

\begin{table}[!b]
\large
\centering
\resizebox{\columnwidth}{!}{%
  \begin{tabular} {ccccc}
  \toprule
    Covid/experiment & Scheduled & Female/Male & Actual & After Denoise\\
\midrule
    Before & 16 & 8/8 & 16 & 11 \\
    During & 6 & 3/3 & 3 & 1 \\ 
\bottomrule
\end{tabular}
 }
\vspace{.1in}
\caption{Comparison of pre/during pandemic data collection}
\label{pandamic_data}
\end{table}

\subsection{Algorithms}
We implement several machine learning algorithms commonly used in the field \cite{breiman1996bagging,breiman2001random,breiman2017classification,chevalier2020statistical} from the scikit-learn \cite{scikit-learn}. For example, Linear Classifiers, Nearest Neighbors, Decision Trees, and Ensemble Methods.  

\textbf{Linear Classifiers}: The Linear Discriminant Analysis(LDA), especially Shrinkage Linear Discriminant Analysis (Shrinkage LDA), and Support Vector Machine(SVM), which works well on high dimensional spaces, perform adequately on EEG datasets based on previous research \cite{lotte2015signal, bashivan2015learning}. These algorithms are simple and computationally efficient because they classify data based on linear functions. In this paper, we use 'LinearDiscriminantAnalsysis' and set the Shrinkage to 'auto' for Shrinkage LDA. We used SVC with RBF and linear kernel functions from the scikit-learn for SVM.

\textbf{Nearest Neighbor}: Such a classifier implements the K-Nearest Neighbor(KNN). The KNN performs pretty well on EEG datasets \cite{lotte2018review}, compared to most of the other classifiers. We use 'KNeighborsClassifier,' and choose k to be five (as there are five tasks). 

\textbf{Decision Tree}: The Decision Trees classifier is easy to understand, implement, and interpret. The computational cost of using the tree is low and can handle both numerical and categorical data. However, it might overfit as trees are too complex. In this paper, we use 'DecisionTreeClassifier.' 

\textbf{Ensemble Methods}: 
We implement Random Forest, Adaboost, and Gradient boosting. Ensemble Machine Learning algorithms combine several weaker learners' predictions and form more robust and accurate predictions. These have been widely used in EEG-based experiments and research \cite{lotte2018review,qu2020multi,qu2020using}. In this paper, we use 'RandomForestClassifier', 'AdaBoostClassifier', and 'GradientBoostingClassifier' from the scikit-learn \cite{scikit-learn}. 

\section{Results}

\subsection{Remote Data Collection}
We scheduled a remote data collection meeting with six end-users during the pandemic, as shown in Table \ref{pandamic_data}. During the zoom meeting, only three participants showed up. Two of them had noise issues. They could not simultaneously get all four electrodes on their skin to generate consistent, stable signals. Only one end-user successfully finished the experiment. We labeled the data from this end-user as data of subject seventeen. 

\subsection{Classification Results}
\begin{table}[!b]
\begin{adjustbox}{width=0.99\columnwidth}
\centering
\begin{tabular} {ccc}
\toprule
   Algorithm & Average Accuracy & Average run-time (s) \\ 
\hline
    Random Forest & \textbf{0.63} & 37.1 \\
    RBF SVM & \textbf{0.61} & 15.2 \\ 
    GradientBoost & 0.58 & 119.0 \\ 
    Nearest Neighbors & 0.55 & \textbf{2.3} \\ 
    Decision Tree & 0.5 & \textbf{2.7}\\ 
    Linear SVM & 0.48 & 11.3\\ 
    Shrinkage LDA & 0.48 & 1.9 \\ 
    LDA & 0.48 & 1.9 \\ 
    Adaboost & 0.44 & 7.1 \\ 
\bottomrule
\end{tabular}
\end{adjustbox}
\vspace{.1in}
\caption{\label{accuracy_runtime} Algorithms with Accuracy and Run-time }
\end{table}

\begin{table}[!b]
\begin{adjustbox}{width=0.98\columnwidth,center}
\begin{tabular} {ccc}
\toprule
   Algorithm & Accuracy(EEG4Students) & Accuracy(\cite{qu2020multi}) \\ 
\midrule
    Random Forest & \textbf{0.63} & \textbf{0.64} \\
    RBF SVM & \textbf{0.61} & 0.55 \\ 
    Nearest Neighbors & 0.55 & 0.58 \\ 
    Shrinkage LDA & 0.48 & 0.54\\ 
    LDA & 0.48 & 0.54 \\ 
    Adaboost & 0.44 & 0.57 \\ 
\bottomrule
\end{tabular}
\end{adjustbox}
\vspace{.1in}
\caption{\label{accuracy_compare_with_paper} Accuracy Comparison between EEG4Students and the TCR\cite{qu2020multi} paper}
\end{table}

We report the average accuracy for all subjects and the runtime of each classifier for each participant in table \ref{accuracy_runtime}. As we can see, Random Forest has the highest accuracy in our experiment, and the SVM with RBF kernel is the second best performing classifier on the dataset. Though Nearest Neighbors does not perform as well as the Random Forest and the RBF SVM, it is one of the fastest algorithms on personal computers. In contrast, although the average accuracy of GradientBoost is 3\% higher than that of the KNN, it is the most time-consuming algorithms we experimented with. The Adaboost performs the worst compared to other algorithms. We present the comparison between the average accuracy of EEG4Students and the TCR paper in Table \ref{accuracy_compare_with_paper}.

\begin{figure}[!b]
\centering
  \includegraphics[width=0.999\columnwidth]{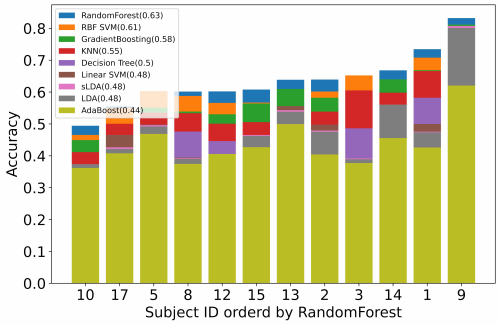}
  \caption{Algorithm performance among different subjects. The X axis is the subject id ordered by the accuracy of the Random Forest classifier, and the y axis is the accuracy. The legend shows the classifiers' names and average accuracy among all participants.}
  \label{fig:CompareAlgorithms}
\end{figure}

\begin{figure}[!b]
\centering
  \includegraphics[width=0.999\columnwidth]{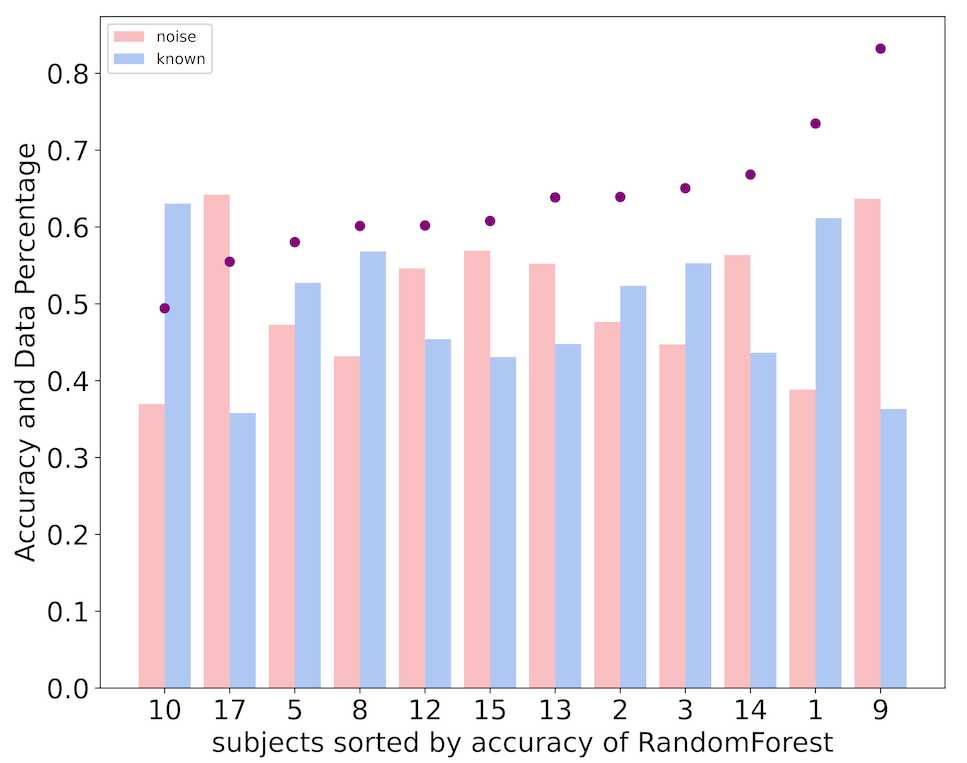}
  \caption{Visualization of data noise and data left to cross-validate for each subject. The X axis is the subject id ordered by the accuracy of the best performing classifier, Random Forest, and the y axis is the percentage. The purple dot represents the corresponding accuracy of the Random Forest.}
  \label{fig:subjectsNoise}
\end{figure}

The individual difference may impact the accuracy of each subject. But we can still recognize a general pattern from Fig \ref{fig:CompareAlgorithms}. We order all twelve subjects by prediction accuracy using Random Forest. Most of the algorithms demonstrate consistent patterns for the different algorithms. Random Forest and SVM with RBF kernel are above most other algorithms. As shown in Fig \ref{fig:subjectsNoise}, noise is also a concern for our data collection and analysis. As we can see, six subjects had more than 50\% noise. We keep the threshold of including the subject to 35\% of the remaining data, as we believe that when a participant has little data left, the high accuracy contributes little to our research.

\begin{figure}[!t]
\centering
  \includegraphics[width=0.999\columnwidth]{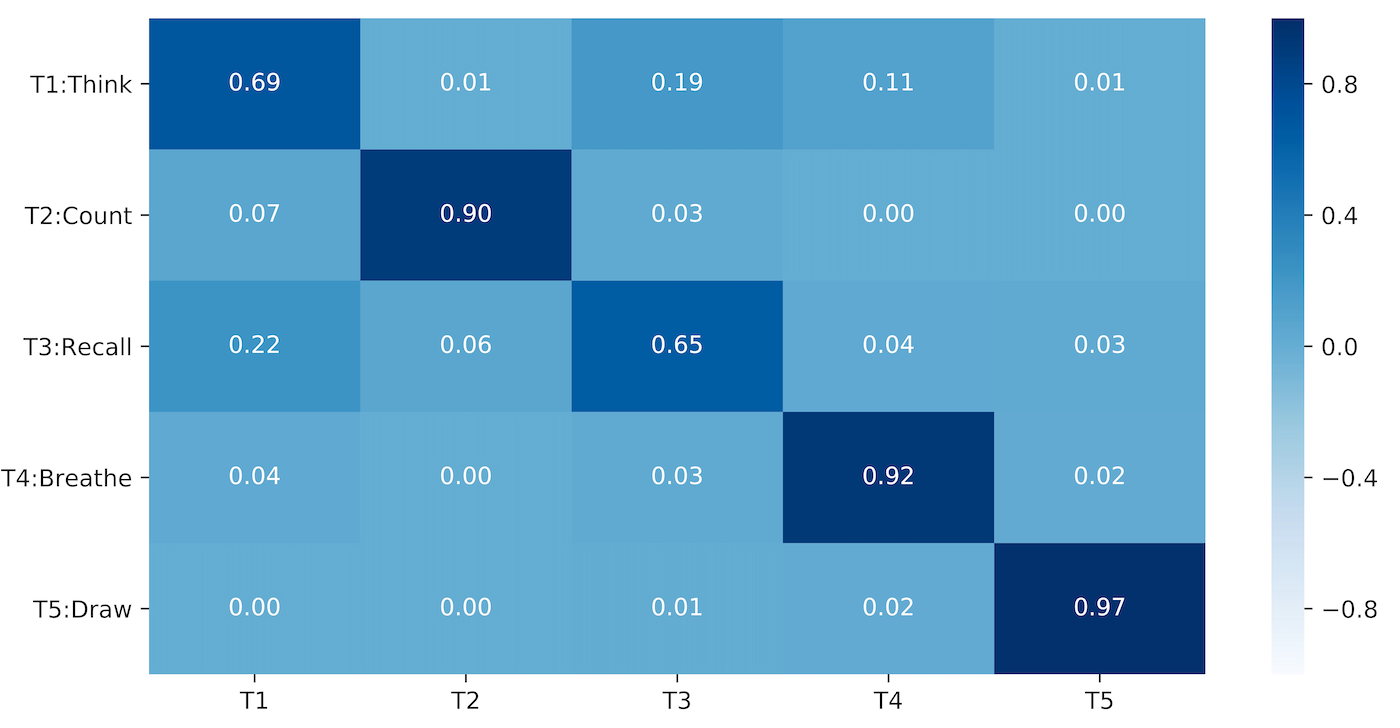}
  \caption{Heatmap for all five tasks of subject 9. The X-Axis is the five designed tasks (Think, Count, Recall, Breathe, and Draw), and the Y-Axis is the predicted tasks.}
  \label{fig:subject_9_RF_heatmap}
\end{figure}

\begin{figure}[!t]
\centering
  \includegraphics[width=0.999\columnwidth]{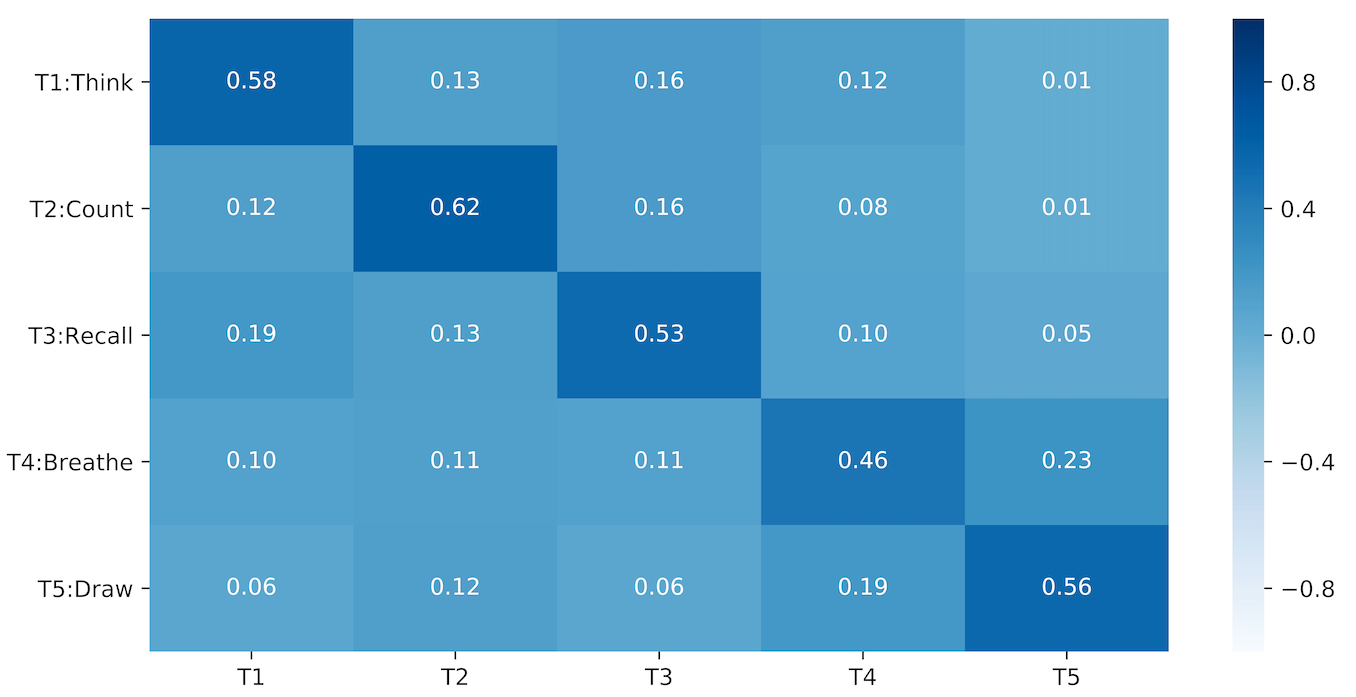}
  \caption{Heatmap for all five tasks of subject 17, the remote participant of our data collection experiment. The X-Axis is the five designed tasks (Think, Count, Recall, Breathe, and Draw), and the Y-Axis is the predicted tasks.}
  \label{fig:subject_17_RF_heatmap}
\end{figure}

Figure \ref{fig:subject_9_RF_heatmap} and Figure \ref{fig:subject_17_RF_heatmap} are heatmaps for subject 9 and subject 17 for all five tasks, respectively. Subject 9 yields the highest Random Forest accuracy among in-person participants. Subject 17 is the only participant we experimented with among virtual participants.
 
As shown in Figure \ref{fig:subject_9_RF_heatmap}, the diagonal is the tasks that have been correctly predicted. The other cells give the team a sense of how one task may be misclassified as other tasks. For subject 9, the Random Forest can predict the tasks Count, Breathe, and Draw with high accuracy, while not so good on Think and Recall. However, as we can see from Figure \ref{fig:subjectsNoise}, subject 9 has a lot of noise compared to other subjects.

Figure \ref{fig:subject_17_RF_heatmap} is the classification results for subject 17, the subject we collected completely in virtual settings. The Random Forest performs well, but not so well compared to subject 9. We also note from Figure \ref{fig:subjectsNoise} that subject 17 has much noise that we exclude for training. 

Such visual feedback also helped the research team and the end-users better understand what task pairs were easy to be confused with each other.

\section{Discussion}
\subsection{Virtual Coaching Challenges}
This paper proposes an approach for non-expert independent researchers to collect EEG-based BCI data with affordable non-clinical devices. When performing test trials with Muse headsets, we provide a general guideline, as shown in figure \ref{fig:data_collection_app} and \ref{fig:data_collection_user_coach}, which shows promising results in many EEG data collections performed by naive EEG end-users. This general guideline demonstrates a decision tree for non-expert researchers to acquire data collection hardware and software. We also present our data collection flow, which forms a closed loop between the researchers and the experimental subjects. In addition, we elaborate on our data-cleaning analysis procedures. 

Significant progress has been made in user training for EEG-based BCI studies, while the EEG4Students serves as a stepping stone for further improved training programs in future research. 

However, some limitations are identified along the course of our project. Our project spans from pre-pandemic to post-pandemic time. We find that in-person data collection trials are significantly more efficient than trials that take place virtually during the global pandemic. If an end-user runs into a distinct situation during Equipment connection or data collection due to personal device usage, the problem can hardly be resolved. For instance, about half of the EEG coaches are issued a newer version of the 2018 Muse headset compared to the original ones from 2016; the updated model is so distinct that it is incompatible with the intended Muselab application. In order to resolve this incompatibility, a new EEG recording application, Mind Monitor, is found, and another protocol is formulated specifically for the 2018 Muse headset. Therefore, a different experiment protocol and data collection have to be developed and delivered to the end-users virtually.

During the pandemic, six subjects (three female and three male participants) were recruited, and each subject was issued a Muse headset and instructions for data collection. The entire coaching and data acquisition process was converted into a virtual established environment. Given the distinct individual problematic data collecting process and the communication barrier set up by the pandemic, multiple participants failed to submit EEG data. As shown in Table \ref{pandamic_data}, in the pre-pandemic BCI experiment that took place in person, all sixteen subjects delivered completed data, and eleven datasets were considered meaningful after denoised. By the end of the study, during the pandemic, three out of six participants participated in the experiment. Only one out of the six participants delivered a completed dataset. More BCI user training, as \cite{roc2020review} suggested, could help solve such issues. Therefore, We ought to explore more strategies and updated methods that could grant us efficiency when data collection has to be completed in a virtual environment.

Even though as much detail and trial and error experience we managed to include in our guideline, there are chances that individual cases develop distinct issues. Our priority is to return to an in-person setting when allowed for data collection efficiency. As we perfect our data collection protocol, we will recruit more subjects and potential EEG coaches to enlarge the current dataset exponentially. Our future work includes developing a study/work community for college students as end-users and EEG coaches. In this interactive community, they can discuss their experience with non-clinical devices' use of EEG-based data collection and possibly establish solutions to various issues after the encounter.

\subsection{Machine Learning Algorithms}

Comparatively low accuracies by Machine Learning and Deep Learning algorithms in medical tasks are common \cite{ordun2022intelligent, semigran2016comparison, razzaki2018comparative, richens2020improving, qu2018personalized, qu2020using}. Especially for EEG classification tasks, Machine Learning techniques have struggled to achieve high accuracies due to low signal-to-noise ratio, the outlier issues, and the "mind wandering" states during the experiments \cite{qu2022eeg4home, lotte2018review, lotte2007review}. We should note that the results of the EEG4Students are similar to the paper we are trying to replicate, based on the table \ref{accuracy_compare_with_paper}. Due to the small datasets, we did not implement any deep learning models that are popular in the field of EEG, such as CNN, RNN, and LSTM \cite{craik2019deep, roy2019deep}. We will try to implement these deep learning algorithms as we collect more data with more subjects. In this section, we still provide the guidelines for choosing the Machine Learning classifiers that achieve relatively high accuracies.

\noindent\textbf{Random Forest: } Random Forest yields the highest accuracy in predicting the tasks, based on Figure \ref{fig:CompareAlgorithms}. Also, the Random Forest is easy to implement on personal computers from the scikit-learn.

\noindent\textbf{SVM: } Support Vector Machine with RBF kernel function should be used instead of the SVM with a linear kernel function, as it performes better for all tested subjects. The average accuracy is close to the Random Forest, and its average accuracy is 13\% higher than the linear SVM. 

\noindent\textbf{KNN: } The K Nearest Neighbors performs adequately. The classification is computed by the majority vote of the nearest neighbors of each data point. The KNN is efficient, as the average runtime of the KNN was 2.3 seconds to perform the timewise cross-validation. 

\noindent\textbf{Boosting: } Boosting classifiers in the paper provide various levels of prediction accuracy in our experiment. The accuracy for Adaboost is ranked the lowest, but the GradientBoost is ranked the third among classifiers with significant runtime. 

We recommend using Random Forest and RBF SVM on personal computers. Also, KNN generates a comparable baseline result with the fastest fast runtime. 

\section{Conclusion}
This paper investigates the data collection for EEG-based BCI to develop larger datasets. We explore the possibility of collecting EEG data from college students with affordable devices. The results demonstrate that the proposed framework could simplify the process and contribute to developing a larger EEG dataset.
Here, we present the EEG4Students model. The results show a reasonable prediction accuracy of Random Forest and RBF SVM and interpretable task types for most end-users. In turn could be a building block toward the future of everyone using non-invasive, wireless, and affordable EEG-based BCI systems, whether for college students or other non-experts, to collect BCI data similar to current smartphone usage.

\bibliographystyle{ACM-Reference-Format}
\bibliography{Jason_Gordon}










\end{document}